%% file: colm2026_conference.tex
\documentclass{article}
\usepackage[final]{colm2026_conference}

\usepackage{microtype}
\usepackage{hyperref}
\usepackage{url}
\usepackage{booktabs}
\usepackage{amsmath,amssymb,amsthm,mathtools}
\usepackage{graphicx}
\usepackage{enumitem}
\usepackage{multirow}
\usepackage{lineno}

\usepackage{booktabs}
\usepackage{multirow}
\usepackage[table]{xcolor}
\usepackage{graphicx}
\usepackage{subcaption}
\usepackage{booktabs}
\usepackage{multirow}
\usepackage{makecell}
\usepackage{threeparttable}
\usepackage{adjustbox}
\usepackage{wrapfig}   % 核心：实现文字环绕
\usepackage{graphicx}  % 插入图片
\usepackage{caption}

\definecolor{darkblue}{rgb}{0, 0, 0.5}
\hypersetup{colorlinks=true, citecolor=darkblue, linkcolor=darkblue, urlcolor=darkblue}

\title{QLPO: Quadrant-weighted Sampling for Length-aware Policy Optimization}
\author{
Siwei Chen$^{1,2}$
\quad
SiQi Chen$^{3}$
\quad
Xupeng Miao$^{1,2}$
\quad
Bin Cui$^{1,2,4}$
\\[2mm]
$^{1}$School of Computer Science, Peking University
\\
$^{2}$Beijing Key Laboratory of Software and Hardware Cooperative\\
\hspace*{1em}Artificial Intelligence Systems, Peking University
\\
$^{3}$Department of Electronic Engineering, Tsinghua University
\\
$^{4}$Institute of Computational Social Science, Peking University (Qingdao)
}
\hypersetup{
  pdftitle={QLPO: Quadrant-weighted Sampling for Length-aware Policy Optimization},
  pdfauthor={Siwei Chen, SiQi Chen, Xupeng Miao, and Bin Cui}
}

\newcommand{\name}{QLPO}
\newcommand{\method}{Quadrant-weighted sampling}

\begin{document}

\ifcolmsubmission
\linenumbers
\fi

\maketitle

\begin{abstract}
\input{main-abstract}

\end{abstract}

\input{main-introduction}
\input{main-preliminaries}

\input{main-method}

\input{main-experiments}

\input{main-result}
\input{main-discussion}
\input{main-related}
\input{main-conclusion}

\bibliography{colm2026_conference}
\bibliographystyle{colm2026_conference}

\appendix
\input{appendix}

\end{document}

%% file: main-abstract.tex
Recent large reasoning models often develop long chain-of-thought responses during reinforcement learning (RL), resulting in high inference latency and deployment cost. Existing methods for response length control typically rely on explicit length penalties or additional control modules, which require careful tuning and may compromise reasoning quality. We propose Quadrant-weighted sampling for Length-aware Policy Optimization (\name), a simple resampling-based variant of GRPO that introduces implicit length control without modifying the reward function. QLPO first over-generates candidate responses and then resamples the training group by preserving the empirical correct/incorrect ratio while favoring short correct responses and long incorrect responses. This reshapes the training distribution and implicitly encourages shorter model outputs. Across models ranging from 1.5B to 32B parameters, including both base models and strong reasoning models, QLPO consistently improves the accuracy--length trade-off. It reduces response length by 30\% to 70\% while preserving reasoning performance. These results suggest that structured resampling provides an effective and robust approach to efficient reasoning.

%% file: main-introduction.tex
\section{Introduction}
\begin{wrapfigure}[16]{c}{0.5\linewidth} 
    \centering
    \includegraphics[width=\linewidth]{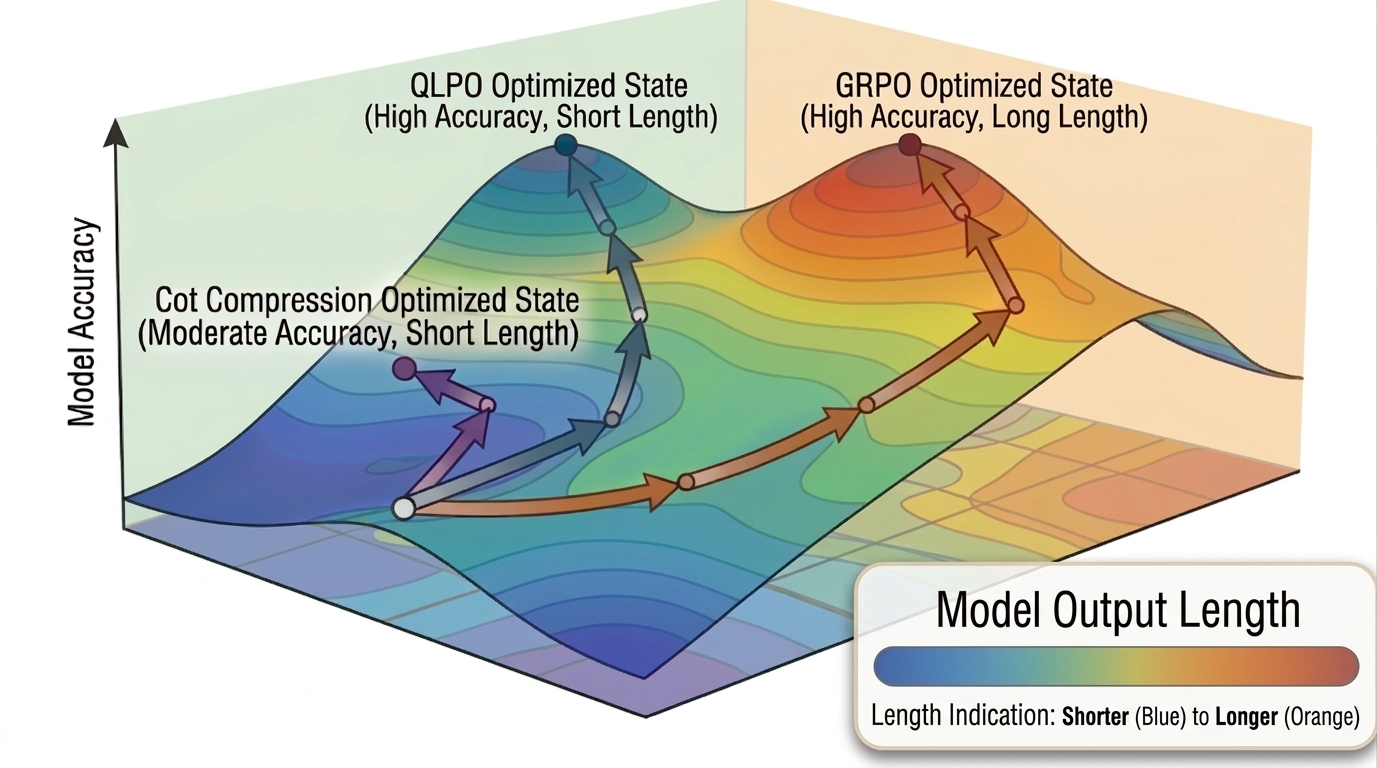}  
    \captionsetup{font=small}  
    \caption{
        Conceptual illustration of the optimization path of different methods in the accuracy-length landscape. 
    }
    \label{fig:qlpo-objective}
\end{wrapfigure}

The integration of Reinforcement Learning (RL) into Large Language Models has catalyzed the emergence of Large Reasoning Models (LRMs), such as OpenAI o-series~\citep{OpenAI-o1,OpenAI-o3}, Gemini 3~\citep{Gemini-3-pro}, Claude 4.5~\citep{Claude-4.5-Sonnet,Claude-4.5-Opus}, Grok 4~\citep{Grok-4}, DeepSeek-R1~\citep{deepseek-r1}, Qwen3~\citep{yang2025qwen3}, Kimi-K2~\citep{kimik2}. RL methods such as GRPO~\citep{deepseekmath-grpo} promote long Chain-of-Thought reasoning, improving self-reflection and allowing LRMs to handle complex mathematical, coding and agentic tasks~\citep{luo2025deepcoder,kimik1.5,feng2025retool}. However, this unbounded optimization paradigm has introduced a severe byproduct: the length explosion phenomenon. Driven by a reward signal that exclusively favors the final correct answer, LRMs consistently develop excessively verbose reasoning paths, generating thousands of redundant tokens. This verbosity not only inflates inference costs to prohibitive levels, but also causes severe deployment latency, creating a critical bottleneck for real-world applications.

\begin{figure}[t]
    \centering
    %\vspace{-12pt}
    \begin{subfigure}[t]{0.48\linewidth}
        \centering
        \includegraphics[width=\linewidth]{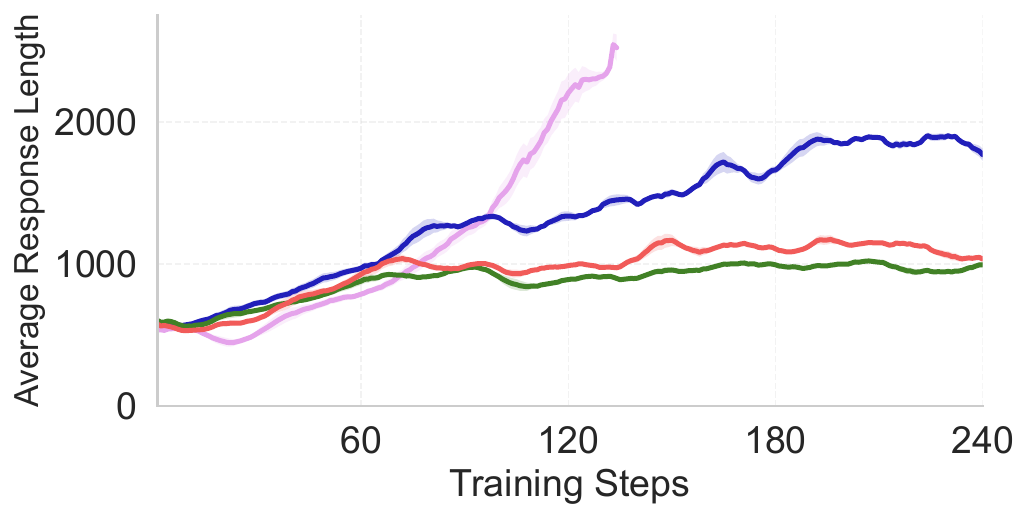}
        %\caption{Average Response Length of different methods.}
        \label{fig:grpo-single}
    \end{subfigure}
    \hfill
    \begin{subfigure}[t]{0.48\linewidth}
        \centering
        \includegraphics[width=\linewidth]{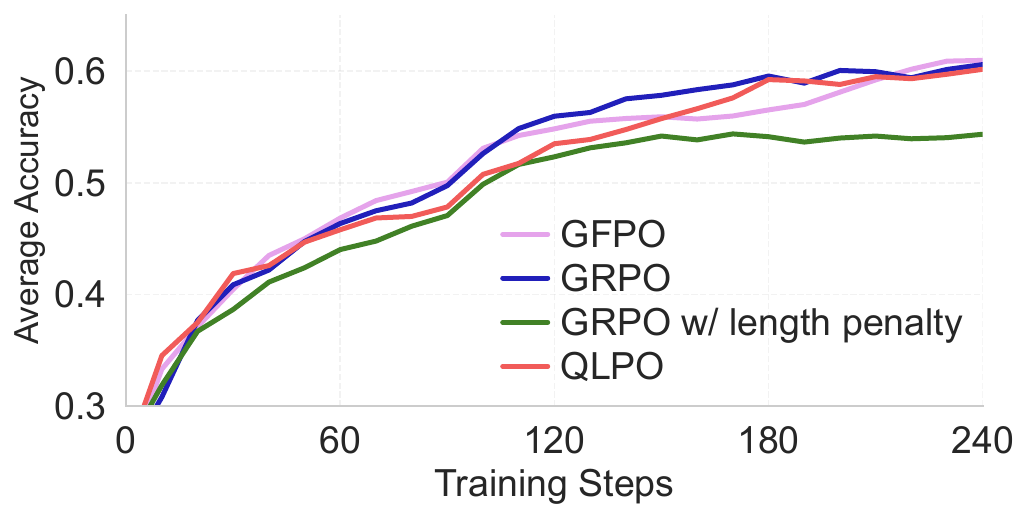}
        %\caption{Average Accuracy of different methods on test set.}
        \label{fig:grpo-multi}
    \end{subfigure}
    
    \caption{Training dynamics of different methods for Qwen2.5-32B on DAPO-MATH.}
    
    %\vspace{-12pt}
    \label{fig:grpo-comparison}
\end{figure}

To enforce conciseness, existing works predominantly rely on explicit length penalties or complex step-level controllers. However, these methods suffer from complex designs with delicate hyperparameters that are difficult to tune. A bad coefficient easily distorts the RL objective, leading to a sharp performance drop on harder problems that require deep and extensive deductive steps. 
Other heuristic resampling approaches, such as Group Filtered Policy Optimization (GFPO) \citep{shrivastava2025gfpo}, often face generalization challenges. Although selecting the shortest responses can be effective in certain scenarios, this strategy ignores the negative gradient contribution of incorrect samples and may even backfire, resulting in worse inference efficiency. Figure \ref{fig:grpo-comparison} illustrates one such failure case, whose underlying mechanism is discussed in detail in Sec.~\ref{gfpo_discussion}.

To alleviate the length explosion problem while overcoming the aforementioned limitations, we shift the paradigm from explicit reward hacking and empirical filtering to structured data resampling. In this paper, we propose \textbf{\name} (\textbf{Q}uadrant-weighted sampling for \textbf{L}ength-aware \textbf{P}olicy \textbf{O}ptimization), a simple and robust RL algorithm. Rather than explicitly penalizing long trajectories in the reward function, QLPO introduces \method\ after the RL rollout phase. Specifically, QLPO dynamically classifies the generated trajectories into four quadrants based on their correctness and relative length. By purposefully sampling specific trajectories before computing the policy gradient, QLPO naturally coerces the model to favor concise reasoning while protecting the model's capacity on complex tasks.
We conduct a comprehensive evaluation of QLPO, demonstrating its effectiveness and robustness. Our primary contributions are summarized as follows:
\begin{itemize}
    
    \item We introduce \textbf{QLPO}, a sampling-based variant of GRPO that achieves length-aware policy optimization while leaving the reward function, advantage estimator, and downstream GRPO optimization unchanged.

    \item We introduce a \textbf{group--trajectory gradient analysis} for group-based RL algorithms, which decomposes optimization into group-level gradient allocation induced by relative reward statistics and trajectory-level token-wise gradient realization. Under this view, QLPO is not a reward-shaping method but a selector that changes group composition and thereby implicitly encourages shorter outputs while preserving the main correctness signal.
    
    \item We show empirically that this simple modification consistently improves the accuracy--length frontier across a broad range of models, from 1.5B to 32B parameters and across both base and strong reasoning models. QLPO achieves substantial reductions in response length while maintaining competitive performance on challenging mathematical and scientific benchmarks.

\end{itemize}

%% file: main-preliminaries.tex
\section{Preliminaries}
\label{preliminaries}
\textbf{Group Relative Policy Optimization} \citep{deepseekmath-grpo} is a widely used reinforcement learning algorithm that removes the need for a critic by estimating advantages from grouped samples \citep{ppo}. Let \(x \sim \mathcal D\) denote a prompt. For each prompt, GRPO samples a group of \(M\) responses \(\mathcal G(x)=\{y_1,\dots,y_M\}\) with \(y_i\sim \pi_{\theta_{\mathrm{old}}}(\cdot\mid x)\), and computes the group-relative normalized advantage
\(
A_i=\frac{r(x,y_i)-\bar r}{\sigma_r}
\),
where \(\bar r=\frac{1}{M}\sum_{j=1}^M r(x,y_j)\) and 
\(\sigma_r=\sqrt{\frac{1}{M}\sum_{j=1}^M (r(x,y_j)-\bar r)^2}\).

The GRPO objective is given by
\begin{equation}
\begin{aligned}
\mathcal L_{\mathrm{GRPO}}(\theta)
&=
\mathbb E_{x,\{y_i\}}
\Bigg[
\frac{1}{\sum_{i=1}^M |y_i|}
\sum_{i=1}^M
\sum_{t=1}^{|y_i|}
\min\!\Big(
\rho_{i,t}(\theta)A_i,\,
\mathrm{clip}\!\big(\rho_{i,t}(\theta),\,1-\epsilon,\,1+\epsilon\big)A_i
\Big)
\Bigg] \\
&\quad
-\beta\, D_{\mathrm{KL}}\!\left(\pi_\theta \,\|\, \pi_{\theta_{\mathrm{ref}}}\right).
\end{aligned}
\label{eq:grpo_loss}
\end{equation}
where \(\rho_{i,t}(\theta)=
\frac{\pi_\theta(y_{i,t}\mid x,y_{i,<t})}
{\pi_{\theta_{\mathrm{old}}}(y_{i,t}\mid x,y_{i,<t})}\),
and \(D_{\mathrm{KL}}\) denotes the KL divergence between the current and ref policies. 
We adopt token-level loss normalization, which is the default choice in VeRL.

\begin{figure}[t]
    %\vspace{-12pt}
    \centering
    \includegraphics[width=\linewidth]{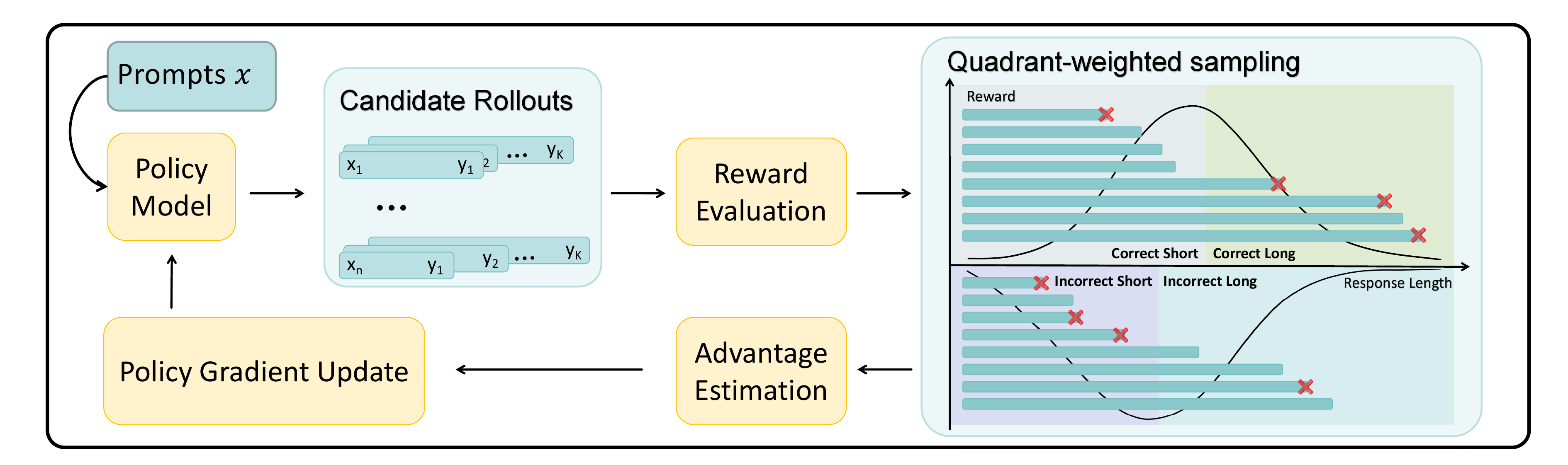}
    \caption{Overview of the QLPO training pipeline. QLPO first generates $K$ candidate trajectories, partitions them according to correctness and relative length, and resamples $M$ trajectories for standard group-relative advantage estimation and policy optimization.}
    %\vspace{-12pt}
    \label{fig:framework}
\end{figure}

%% file: main-method.tex
\section{Quadrant-weighted sampling for Length-aware Policy
Optimization}
\subsection{Overall pipeline}
QLPO is a resampling-based variant of GRPO.
For each prompt \(x\), instead of directly applying GRPO to a group of size \(M\), we first sample an over-generated candidate pool
\[
\widetilde{\mathcal G}(x)=\{y_1,\dots,y_K\}, \qquad K>M,
\]
from the rollout policy \(\pi_{\theta_{\mathrm{old}}}\), and then construct a resampled group of size \(M\) for a standard GRPO update. We first partition the candidate pool by correctness, where \(c(x,y)\in\{0,1\}\) is the correctness indicator: \(\widetilde{\mathcal G}^{+}(x)=\{y\in\widetilde{\mathcal G}(x): c(x,y)=1\}\) and \(\widetilde{\mathcal G}^{-}(x)=\{y\in\widetilde{\mathcal G}(x): c(x,y)=0\}\). Let \(\widetilde N^{+}(x)=|\widetilde{\mathcal G}^{+}(x)|\) and \(\widetilde N^{-}(x)=|\widetilde{\mathcal G}^{-}(x)|\), with \(\widetilde N^{+}(x)+\widetilde N^{-}(x)=K\). We then set the target positive and negative counts in the final group as \(N^{+}(x)=\mathrm{round}\!\left(M\cdot \frac{\widetilde N^{+}(x)}{K}\right)\) and \(N^{-}(x)=M-N^{+}(x)\), so that the final group contains exactly \(M\) samples while approximately preserving the empirical correct/incorrect ratio of the over-generated pool.

QLPO next partitions each correctness class into short and long responses within the same prompt. For each prompt \(x\), we sort responses in \(\widetilde{\mathcal G}^{+}(x)\) by length and split them into two halves,
forming \(\widetilde{\mathcal G}^{+}_{\mathrm{short}}(x)\) and \(\widetilde{\mathcal G}^{+}_{\mathrm{long}}(x)\).
We do the same for \(\widetilde{\mathcal G}^{-}(x)\), obtaining \(\widetilde{\mathcal G}^{-}_{\mathrm{short}}(x)\) and \(\widetilde{\mathcal G}^{-}_{\mathrm{long}}(x)\).
When a class has odd cardinality, the two subsets differ by at most one sample.

We then perform quadrant-weighted sampling to introduce a length preference. A length preference coefficient
\(\alpha \in (0,1)\) specifies the desired within-class retention ratio:
\[
\text{Correct Long}:\text{Correct Short}=\alpha:1,
\qquad
\text{Incorrect Short}:\text{Incorrect Long}=\alpha:1.
\]
Thus, when \(\alpha<1\), QLPO favors short-correct and long-incorrect responses. When
\(\alpha=1\), this length preference disappears, reducing QLPO to GRPO. We prioritize the preferred quadrants, namely \((\text{Correct}, \text{Short})\) and \((\text{Incorrect}, \text{Long})\). So in practice, we use the ceiling function to determine the target numbers of samples drawn from these two quadrants, and obtain the sample counts for the remaining two quadrants by subtraction. This ensures that the final resampled group always contains exactly \(N^{+}(x)\) correct samples and \(N^{-}(x)\) incorrect samples. Accordingly, the target counts are calculated as
\[
\begin{aligned}
\widehat N^{+}_{\mathrm{short}}(x) &= \Bigl\lceil \frac{1}{1+\alpha}N^{+}(x) \Bigr\rceil, \qquad
\widehat N^{+}_{\mathrm{long}}(x) = N^{+}(x)-\widehat N^{+}_{\mathrm{short}}(x),\\
\widehat N^{-}_{\mathrm{long}}(x) &= \Bigl\lceil \frac{1}{1+\alpha}N^{-}(x) \Bigr\rceil, \qquad
\widehat N^{-}_{\mathrm{short}}(x) = N^{-}(x)-\widehat N^{-}_{\mathrm{long}}(x).
\end{aligned}
\]

Finally, we sample trajectories uniformly within each quadrant to form the \(M\) responses for subsequent policy update. The general pipeline of QLPO is demonstrated in Figure \ref{fig:framework}.

\subsection{A Group--Trajectory Gradient View of GRPO and QLPO}

We propose a group- and trajectory-level gradient view to analyze group-based RL algorithms. The key idea is to decompose the update into two coupled levels: \textbf{group-level gradient allocation} and \textbf{trajectory-level token realization}. At the group level, the composition of the selected group determines how the gradient mass is distributed across trajectories through the group-relative advantages. At the trajectory level, this allocated signal is realized as token-wise gradients accumulated along each selected response. We follow the notation introduced in Section \ref{preliminaries}. Let \(G(x)\subseteq \widetilde{\mathcal G}(x)\) denote the final selected group used for the GRPO update. \(\bar r\) sets the center of the update, while \(r(x,y_i)-\bar r\) determines whether a trajectory is reinforced or suppressed relative to the other samples in the same group.

\paragraph{Group-level gradient allocation.}
For a fixed prompt \(x\) and selected group \(G(x)\), the GRPO gradient can be viewed as
\[
\nabla_\theta \mathcal L_{\mathrm{GRPO}}(\theta\mid x,G)
=
\sum_{y\in G(x)} A(y)\,\psi(y;\theta)
-\beta \nabla_\theta D_{\mathrm{KL}}(\pi_\theta\|\pi_{\theta_{\mathrm{ref}}}),
\]
where \(A(y)\) is the group-relative advantage associated with trajectory \(y\in G(x)\), and \(\psi(y;\theta)\) denotes the trajectory-level gradient realization. For simplicity, we ignore the clipping-induced case distinction and use this decomposition only to expose the two-level structure of the update.
Hence, the sampled group determines optimization through a group-level weighting rule: the average reward defines the baseline, and the normalized deviation from that baseline determines the sign and strength of each trajectory's contribution. Thus, GRPO allocates update strength according to each trajectory’s relative position within the group: positive advantages reinforce a trajectory, negative advantages suppress it, and the magnitude of the advantage determines the strength of the effect.

\paragraph{Trajectory-level gradient realization.}
Let \(\mathcal T(G)\) denote the set of valid generated tokens in \(G(x)\), and let \(N_G:=|\mathcal T(G)|\). For a trajectory \(y\in G(x)\), define its token-accumulated gradient contribution by
\[
\psi(y;\theta)
:=
\frac{1}{N_G}\sum_{t=1}^{|y|}
\nabla_\theta \widetilde{\ell}_t(y;\theta),
\]
where \(\widetilde{\ell}_t(y;\theta)\) denotes the token-level clipped PPO term with the trajectory-level advantage factored out. Then the GRPO gradient for a fixed selected group can be written as
\[
\nabla_\theta \mathcal L_{\mathrm{GRPO}}(\theta\mid x,G)
=
\sum_{y\in G(x)} A(y)\,
\frac{1}{N_G}\sum_{t=1}^{|y|}
\nabla_\theta \widetilde{\ell}_t(y;\theta)
-\beta \nabla_\theta D_{\mathrm{KL}}(\pi_\theta\|\pi_{\theta_{\mathrm{ref}}}).
\]
This expression makes the two-level optimization structure explicit. At the group level, the group-relative advantage \(A(y)\) determines the sign and relative magnitude of each trajectory's contribution, and therefore sets the dominant correctness-oriented direction of the update. In this sense, it is the group-level advantage that carries the main signal associated with accuracy and reasoning quality. At the trajectory level, the token-wise gradients along \(y\) determine how this group-level signal is realized in parameter space through token-level accumulation. The resulting update is therefore jointly determined by group-level gradient allocation and trajectory-level gradient realization: the former governs the main performance-relevant direction of optimization, while the latter determines how that direction is concretely expressed through aggregated token gradients.

Under this view, QLPO introduces a selection preference over trajectories rather than modifying the reward or advantage itself. By preferentially retaining short-correct and long-incorrect trajectories, it changes which trajectories contribute gradient mass to the realized update, without directly perturbing the correctness signal encoded in the reward. This is importantly different from explicit length penalties: once a penalty term is added to the reward, the group-relative advantages are recomputed from the penalized rewards, which directly alters group-level gradient allocation. As a result, a long but correct trajectory that would have received a positive advantage under the original reward can become weakly positive or even negative after penalization. In this sense, length penalty does not merely discourage verbosity; it can change both the magnitude and even the sign of the optimization signal at the group level. By contrast, QLPO preserves the dominant correctness-oriented signal carried by the original group-relative advantage, while biasing the realized token-level update toward shorter successful responses and away from longer unsuccessful ones. Repeated over training, this induces an implicit preference for shorter outputs without directly corrupting the reward-defined objective.

\subsection{GFPO from the Group--Trajectory Gradient View}
\label{gfpo_discussion}

GFPO \citep{shrivastava2025gfpo} is closely related to QLPO, but under the group--trajectory gradient view it has two structural weaknesses. First, a shortest-only selector does not preserve the correct/incorrect composition of the candidate pool, and thus perturbs the group-level gradient allocation. Second, shortest-only filtering does not guarantee shorter outputs. Under token-level loss normalization, the update depends on advantage-weighted token gradients aggregated over the selected group. For a selected group \(G\), define the token-weighted average advantage
$
\bar A_{\mathrm{tok}}(x;G)
:=
\frac{1}{\sum_{y\in G}|y|}
\sum_{y\in G}|y|\,A(y).
$
Even in a relatively short-response region, \(\bar A_{\mathrm{tok}}(x;G)\) can be negative when negative-advantage trajectories contribute more token mass than positive-advantage ones. In that case, the update suppresses probability mass in the currently selected short region, causing training to drift away from it and worsening length explosion. We observe this behavior for GFPO in our setup (Figure~\ref{fig:grpo-comparison}). By contrast, QLPO preserves the group-level correctness signal and only reweights trajectories within each class, making it more reliable.

%% file: main-experiments.tex
\section{Experiments}

\subsection{Experimental Setup}

\textbf{Models and datasets.}
We evaluate QLPO on a diverse set of models, including Qwen2.5-3B, Qwen2.5-32B, Qwen3-30B-A3B, DeepScaleR-1.5B-Preview (DeepScaleR-1.5B) and DeepSeek-R1-Distill-Qwen-7B (DS-7B)~\citep{yang2024qwen25math,yang2025qwen3}. 
These models cover both dense and MoE architectures, as well as different levels of reasoning ability, allowing us to test whether QLPO generalizes across model families and capability regimes.
We use MATH-lighteval~\citep{hendrycks2021measuringmathematicalproblemsolving} for the 3B setting and DAPO-MATH~\citep{DAPO} for the remaining models.
In addition to text reasoning, we also conduct preliminary experiments in multimodal reasoning and code generation. Specifically, we evaluate QLPO for Qwen2.5-3B-VL~\citep{bai2025qwen25vl} on Geo3K~\citep{lu2021intergps} and for Phi-3.5-mini-instruct~\citep{abdin2024phi3} on Eurus-2-RL-Data~\citep{cui2025process}, to assess whether our method transfers beyond standard text-only reasoning settings.

%\textbf{Baseline.} We compare QLPO with standard GRPO trained model. We also evaluate a lot of open-source models like L1, Laser and Thinkprune.
\paragraph{Baselines.}
Our primary baseline is standard GRPO with an update-group size of \(M=8\). QLPO samples a candidate pool of \(K=16\) trajectories and selects \(M=8\) trajectories for the policy update. Thus, standard GRPO and QLPO use the same update-group size but different rollout budgets. To isolate the effect of quadrant-weighted selection from the effect of additional sampling, we further evaluate rollout-matched GRPO with \(K=M=16\), which generates the same number
of trajectories as QLPO but updates the policy using all of them. We also compare against GFPO with \(K=16\) and \(M=8\).

We additionally compare against several open-source efficiency-oriented reasoning models, including L1, Laser, and ThinkPrune, using their publicly released checkpoints \citep{aggarwal2025l1,liu2025laser,fang2025thinkless,DBLP:journals/corr/abs-2504-01296}.

\textbf{Training Configuration and Hyperparameters.} 
We implement  QLPO based on VeRL framework~\citep{sheng2025hybridflow} with only about 200 lines of additional Python code. 
Unless otherwise specified, all methods are trained on 32 NVIDIA H20 GPUs with a global batch size of 128 using the Adam optimizer with a learning rate of \(1\times10^{-7}\). For standard GRPO, we use a group size of \(M=8\). In QLPO, we first sample an over-generated candidate pool with size \(K=16\), apply quadrant-aware resampling, and then retain \(M=8\) responses for advantage estimation and policy updates. Therefore, QLPO has the same policy-update group size as standard GRPO, but
uses a larger rollout candidate pool. The rollout-matched comparison in Section~\ref{sec:rollout_matched} controls for this difference in sampling budget.

\textbf{Evaluation.} 
We evaluate the trained models on five widely used reasoning benchmarks: MATH-500~\citep{DBLP:journals/corr/abs-2305-20050}, GSM8K~\citep{DBLP:journals/corr/abs-2110-14168}, OlympiadBench~\citep{DBLP:journals/corr/abs-2402-14008}, GPQA~\citep{rein2024gpqa} and AIME 2024. Unless otherwise specified, we sample model responses at temperature \(T=0.8\) with a maximum generation length of 32k tokens. We generate 16 responses per prompt for AIME24 and 3 responses per prompt for the remaining benchmarks. For GSM8K and MATH-500, correctness is determined by structured answer extraction followed by normalized exact matching. For the remaining benchmarks, we first apply rule-based matching; only when such matching fails do we fall back to GPT-4.1 as an auxiliary verifier. We report pass@1 accuracy and average output length, and use these metrics to evaluate the accuracy-efficiency trade-off of each method.

%% file: main-result.tex
\subsection{Main Results}

\begin{table*}[t]
%\vspace{-12pt}
\centering
\small
\renewcommand{\arraystretch}{1.2}
\setlength{\tabcolsep}{4.5pt}
\begin{tabular}{l cc cc cc cc cc}
\toprule
\multirow{2}{*}{Model} & \multicolumn{2}{c}{GSM8K} & \multicolumn{2}{c}{MATH-500} & \multicolumn{2}{c}{OlympiadBench} & \multicolumn{2}{c}{GPQA} & \multicolumn{2}{c}{AIME24} \\
\cmidrule(lr){2-3} \cmidrule(lr){4-5} \cmidrule(lr){6-7} \cmidrule(lr){8-9} \cmidrule(lr){10-11}
& Acc. & Len. & Acc. & Len. & Acc. & Len. & Acc. & Len. & Acc. & Len. \\
\midrule

%\multicolumn{11}{l}{\textit{\textbf{1.5B Models}}} \\
DeepScaleR-1.5B & 77.3 & 4,021 & 82.4 & 4,159 & 46.6 & 9,265 & 17.2 & 15,062 & 38.8 & 9,978 \\
+ GRPO & 85.2 & 3,434 & 87.2 & 4,095 & 50.7 & 17,138 & 28.2 & 27,758 & 40.8 & 9,674 \\
\rowcolor{gray!10}
+ QLPO & 85.4 & 1,707 & 87.8 & 2,639 & 50.0 & 6,621 & 28.2 & 23,199 & 40.8 & 6,285 \\
\addlinespace

%\multicolumn{11}{l}{\textit{\textbf{7B Models}}} \\
DS-7B & 89.3 & 2,145 & 88.0 & 4,174 & 54.6 & 8,977 & 38.4 & 11,321 & 56.7 & 13,144 \\
+ GRPO & 91.2 & 2,388 & 92.1 & 4,119 & 58.5 & 8,213 & 44.4 & 19,321 & 57.1 & 12,237 \\
\rowcolor{gray!10}
+ QLPO & 92.3 & 1,176 & 92.2 & 2,780 & 58.7 & 6,449 & 44.0 & 14,135 & 57.9 & 9,223 \\
\addlinespace

%\multicolumn{11}{l}{\textit{\textbf{3B Models}}} \\
Qwen2.5-3B & 76.5 & 291 & 56.2 & 1,444 & 20.5 & 4,000 & 28.8 & 1,798 & 4.5 & 4,798 \\
+ GRPO & 86.3 & 414 & 67.6 & 1,546 & 23.1 & 1,309 & 32.8 & 2,225 & 5.0 & 4,982 \\
\rowcolor{gray!10}
+ QLPO & 84.8 & 300 & 69.4 & 1,013 & 26.1 & 885 & 31.7 & 721 & 5.0 & 2,437 \\
\addlinespace

%\multicolumn{11}{l}{\textit{\textbf{32B Models}}} \\
Qwen2.5-32B & 90.6 & 921 & 67.8 & 2,574 & 29.0 & 2,943 & 35.8 & 1,082 & 8.7 & 2,909 \\
+ GRPO & 96.0 & 341 & 87.6 & 862 & 45.2 & 1,397 & 54.0 & 1,455 & 28.3 & 1,880 \\
\rowcolor{gray!10}
+ QLPO & 95.6 & 295 & 86.2 & 675 & 45.4 & 867 & 56.5 & 978 & 29.1 & 1,446 \\
\addlinespace

%\multicolumn{11}{l}{\textit{\textbf{MoE Models}}} \\
Qwen3-30B-A3B & 96.0 & 1,833 & 96.2 & 5,448 & 88.2 & 10,219 & 56.1 & 6,064 & 78.3 & 14,167 \\
+ GRPO & 96.2 & 1,708 & 97.4 & 4,737 & 90.8 & 9,482 & 59.1 & 4,586 & 83.3 & 13,355 \\
\rowcolor{gray!10}
+ QLPO & 96.1 & 1,215 & 96.8 & 3,528 & 89.3 & 7,070 & 60.6 & 4,016 & 82.1 & 9,692 \\
\bottomrule
\end{tabular}
\caption{Accuracy--length trade-off across models under the base, GRPO, and QLPO settings.}
\label{tab:main_controlled}
\end{table*}

\paragraph{\name\ Improves the Accuracy–Length Trade-off} The magnitude of compression is substantial across a wide range of scales. For example, QLPO reduces OlympiadBench length from 17,138 to 6,621 for DeepScaleR-1.5B, AIME length from 12,237 to 9,223 for DeepSeek-R1-Distill-Qwen-7B and OlympiadBench length from 1,397 to 867 for Qwen2.5-32B. The largest relative reduction on these hard benchmarks reaches 67.6\%, achieved on Qwen2.5-3B over GPQA (2,225 to 721), while the largest absolute reduction reaches 10,517 tokens on DeepScaleR-1.5B over OlympiadBench (17,138 to 6,621).

A central concern for QLPO is whether shorter responses come at the cost of reasoning performance on benchmarks that genuinely require extended multi-step reasoning. Our results show that this is not the case for QLPO. On the hardest benchmarks in Table~\ref{tab:main_controlled}, namely OlympiadBench, GPQA, and AIME24, QLPO matches or even slightly exceeds GRPO in most cases while producing substantially shorter outputs. Across the 15 model--benchmark pairs on these three benchmarks, the few observed drops are all small in magnitude and are well within the range of ordinary training stochasticity rather than evidence of systematic degradation. Overall, these results show that QLPO substantially improves response efficiency without sacrificing, and in some cases even improving, reasoning performance.

\begin{figure}[t]  
    \centering   
    %\vspace{-12pt}
    \includegraphics[width=1.0\linewidth]{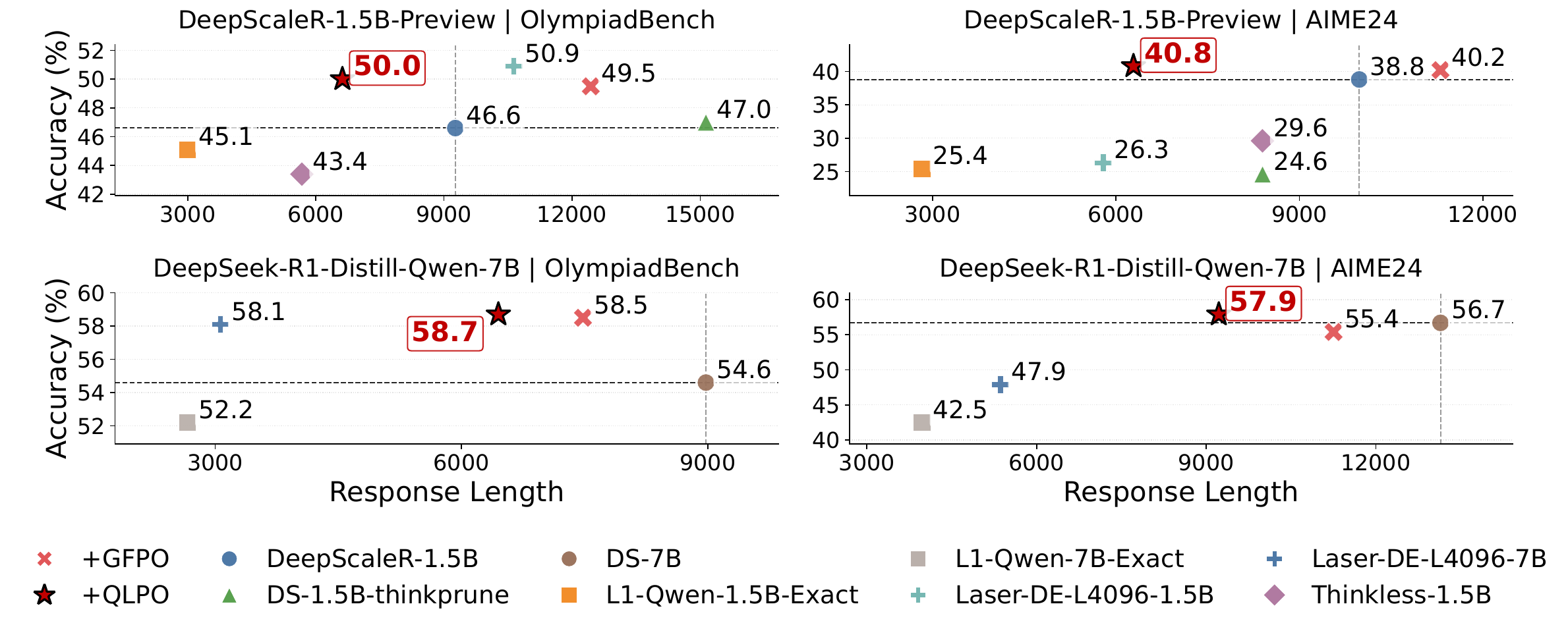}
    %\caption{Comparison with other compression baselines.}
    \caption{Comparison with representative compression baselines. Each point shows a different model, where lower response length and higher accuracy are preferred. Across both 1.5B and 7B settings, QLPO consistently achieves a more favorable trade-off, delivering substantial length reduction while preserving strong reasoning performance.}
    %\vspace{-12pt}
    \label{fig:baseline}
\end{figure}

\paragraph{Accuracy--Length Trade-off Against Other Compression Methods}
Figure~\ref{fig:baseline} compares QLPO with representative compression baselines. While some methods can shorten model outputs more aggressively, they typically do so by imposing a stronger compression bias, which may lead to noticeable degradation in reasoning performance, especially on more challenging benchmarks. By contrast, the main advantage of QLPO is not to maximize raw compression, but to improve reasoning efficiency while better preserving the performance gains of RL post-training. Empirically, in the 1.5B setting, QLPO achieves the strongest AIME24 performance (40.8) among the compared methods, while remaining substantially shorter than the original GRPO policy. On OlympiadBench, it maintains competitive accuracy with a much shorter response length than standard GRPO. In the 7B setting, QLPO again delivers the strongest AIME24 result (57.9) and OlympiadBench result (58.7) despite using fewer tokens than standard GRPO. Taken together, these results suggest that QLPO offers a more reliable accuracy--length trade-off than representative compression baselines.

%\paragraph{\name\ Remains Robust Across Model Scales and Architectures}
%Another notable result is the robustness of QLPO across diverse training setups. Our method performs well on reasoning models, dense base models and MoE models. Across all of these models, QLPO consistently reduces output length relative to GRPO and does not impact downstream reasoning performance. This robustness suggests that QLPO captures a general improvement to group-based RL. By reshaping the within-group training distribution instead of directly penalizing long responses, QLPO can encourage concise reasoning while preserving the correctness signal that makes GRPO effective. Taken together, the results indicate that QLPO is a simple and broadly applicable method for improving reasoning efficiency.

\subsection{\name\ Generalizes Across Diverse Tasks and Model Settings}
Another notable result is the strong and consistent performance of QLPO across a range of text-based reasoning tasks. Across five models, QLPO stably reduces output length while maintaining competitive reasoning performance, demonstrating its effectiveness in improving the accuracy–efficiency trade-off. Beyond standard text reasoning, we also conduct preliminary experiments in multimodal and code generation settings. Specifically, we evaluate QLPO on Qwen2.5-3B-VL with Geo3K and on Phi-3.5-mini-instruct with Eurus-2-RL-Data. In both cases, QLPO achieves clear length reduction while preserving task performance, with around 42\% and 25\% shorter outputs respectively. These results suggest that QLPO is not limited to text-only reasoning, but can generalize to broader reasoning settings, providing a simple and effective approach for improving inference efficiency. We demonstrate the detailed training dynamics in Appendix \ref{append-dynamics}.

\subsection{Training–Inference Trade-off}
The additional training overhead introduced by QLPO is limited in practice. The main source of extra cost lies in rollout, which is primarily memory-bound rather than compute-bound. Because decoding is autoregressive, throughput is constrained by sequential token generation and is often dominated by the long tail of exceptionally long responses. Consequently, increasing the number of rollouts does increase decoding cost, but the resulting wall-clock overhead is less than proportional to the increase in rollout count. Moreover, as training progresses, the model tends to produce shorter responses, which correspondingly reduces both rollout latency and policy-update time greatly. 
Empirically, QLPO incurs at most a 16.3\% increase in wall-clock training time relative to GRPO, and can in some cases be up to 8.4\% faster (Table \ref{tab:appendix_time_length}).
This trade-off is acceptable: a modest increase in training-time cost can be exchanged for substantial savings in inference-time computation.

\begin{figure}[t] 
    \centering 
    %\vspace{-12pt}
    \includegraphics[width=1.0\linewidth]{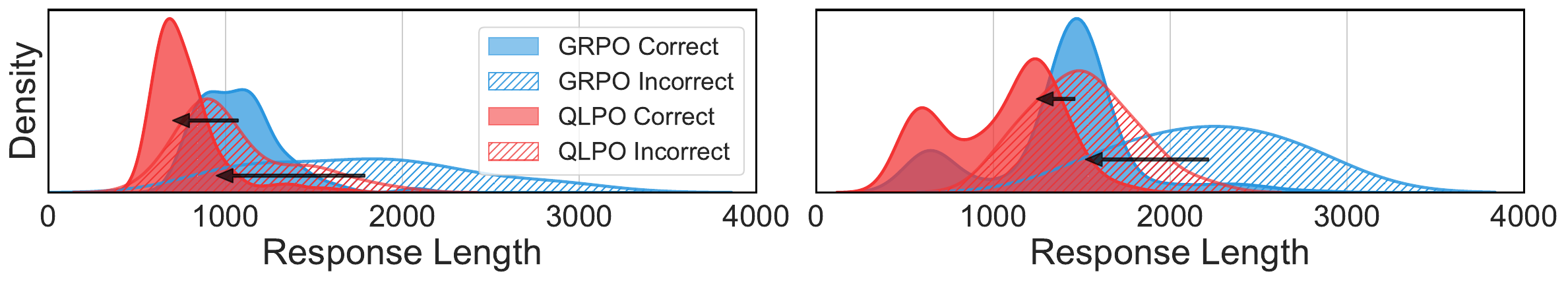}
    \caption{Response Length Distribution Shift Across Trajectory Correctness for DeepScaleR-1.5B-Preview on GSM8K (left) and DAPO-MATH (right). QLPO induces a larger leftward shift for incorrect trajectories.}
    %\vspace{-12pt}
    \label{fig:acc-len}
\end{figure}

\section{Analysis}
\subsection{Rollout-Budget-Matched Comparison}
\label{sec:rollout_matched}

QLPO uses a larger candidate pool than standard GRPO. To disentangle the
benefit of quadrant-weighted selection from the benefit of observing additional
trajectories, we introduce a rollout-budget-matched GRPO baseline. In this
comparison, both QLPO and rollout-matched GRPO generate and evaluate
\(K=16\) trajectories for each prompt. Rollout-matched GRPO uses all 16
trajectories for advantage estimation and policy updates, whereas QLPO selects
\(M=8\) trajectories through quadrant-weighted sampling.

\begin{table}[t]
\centering
\small
\setlength{\tabcolsep}{4.5pt}
\begin{tabular}{llccc}
\toprule
Model & Method & Avg.\ Acc. & Avg.\ Len. & Train Time \\
\midrule
\multirow{3}{*}{DeepScaleR-1.5B}
& GRPO (\(K=8,M=8\))
& 58.4 & 12,419 & 1.00 \\
& GRPO (\(K=16,M=16\))
& 58.2 & 14,734 & 2.13 \\
& QLPO (\(K=16,M=8\))
& 58.4 & 8,090 & 1.04 \\
\midrule
\multirow{3}{*}{Qwen2.5-3B}
& GRPO (\(K=8,M=8\))
& 43.0 & 2,095 & 1.00 \\
& GRPO (\(K=16,M=16\))
& 43.3 & 2,315 & 1.74 \\
& QLPO (\(K=16,M=8\))
& 43.4 & 1,071 & 1.13 \\
\bottomrule
\end{tabular}
\caption{
Rollout-budget-matched comparison. QLPO and 16-rollout GRPO generate
and evaluate the same number of trajectories. QLPO performs policy updates
only on the selected subset of eight trajectories, whereas rollout-matched
GRPO updates on all 16. Accuracy and response length are averaged over the
five main benchmarks. Training time is normalized by standard 8-rollout GRPO
for each model.
}
\label{tab:rollout_matched}
\end{table}

As shown in Table~\ref{tab:rollout_matched}, simply increasing the GRPO group size from 8 to 16 does not improve response efficiency. On both models, rollout-matched GRPO produces slightly longer responses than standard GRPO and incurs substantially higher training cost. In contrast, QLPO observes the same 16-trajectory candidate pool while maintaining comparable accuracy and producing much shorter responses.

Relative to rollout-matched GRPO, QLPO reduces average response length by 45.1\% on DeepScaleR-1.5B and 53.7\% on Qwen2.5-3B. Moreover, because only eight selected trajectories enter advantage estimation and back-propagation, QLPO requires substantially less policy-update computation than 16-rollout GRPO. These results demonstrate that the accuracy--length gain does not arise from access to additional rollouts alone, but from the proposed
structured selection mechanism.

\subsection{Ablation Study}
\label{ablation}

\begin{wrapfigure}[16]{c}{0.5\linewidth}  
    \centering
    \includegraphics[width=\linewidth]{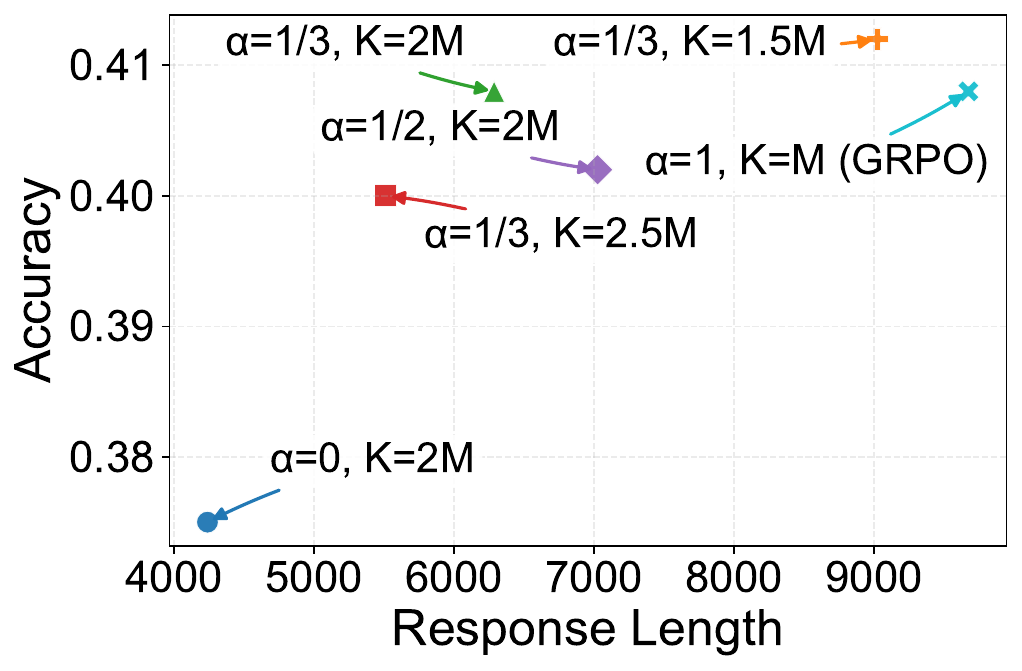} 
    \captionsetup{font=small}  
    \caption{
        %Ablation Study of hyperparameters. 
        Ablation of hyperparameters for DeepScaleR-1.5B-Preview on DAPO-MATH.
    }
    \label{fig:ablation}
\end{wrapfigure}
The effect of QLPO depends on both the candidate pool size $K$ and the length preference coefficient $\alpha$. We conduct an ablation study on DeepScaleR-1.5B-Preview. As shown in Figure \ref{fig:ablation}, for the candidate pool size, we observe that QLPO requires sufficient over-generation to expose useful diversity among sampled responses. When $K=1.5M$, the gain is limited, indicating that the sampled pool is too narrow for quadrant-aware resampling to identify concise high-quality trajectories. In contrast, $K=2M$ already delivers strong and stable improvements, and further increasing $K$ brings only marginal additional benefit.
For the length preference coefficient, $\alpha=1$ reduces QLPO to standard GRPO, since all correct responses are treated equally regardless of length. Empirically, we find that $\alpha=\frac{1}{3}$ works well and leads to a significant reduction in response length in our experiments. By contrast, $\alpha=0$ corresponds to an extreme corner case that completely removes long but correct responses from the selected group, which is overly aggressive. In practice, this may discard informative long-horizon reasoning trajectories and lead to performance degradation. 

%% file: main-discussion.tex
\subsection{Distribution Shift Analysis}

QLPO improves reasoning efficiency by increasing the relative contribution of long incorrect responses, which makes it particularly effective at suppressing verbose yet unproductive reasoning trajectories. At the same time, QLPO still retains a portion of long correct responses, so the model’s ability to learn useful long-horizon reasoning paths is preserved rather than removed. As shown in Figure~\ref{fig:acc-len}, both correct and incorrect response distributions shift toward shorter lengths under QLPO, but the shift is visibly larger for incorrect responses. On GSM8K and DAPO-MATH, the average response length of incorrect trajectories decreases by 38\%, while that of correct trajectories decreases by 23\%. 

\begin{figure}[t]  
    \centering   
    %\vspace{-12pt}
    \includegraphics[width=1.0\linewidth]
    {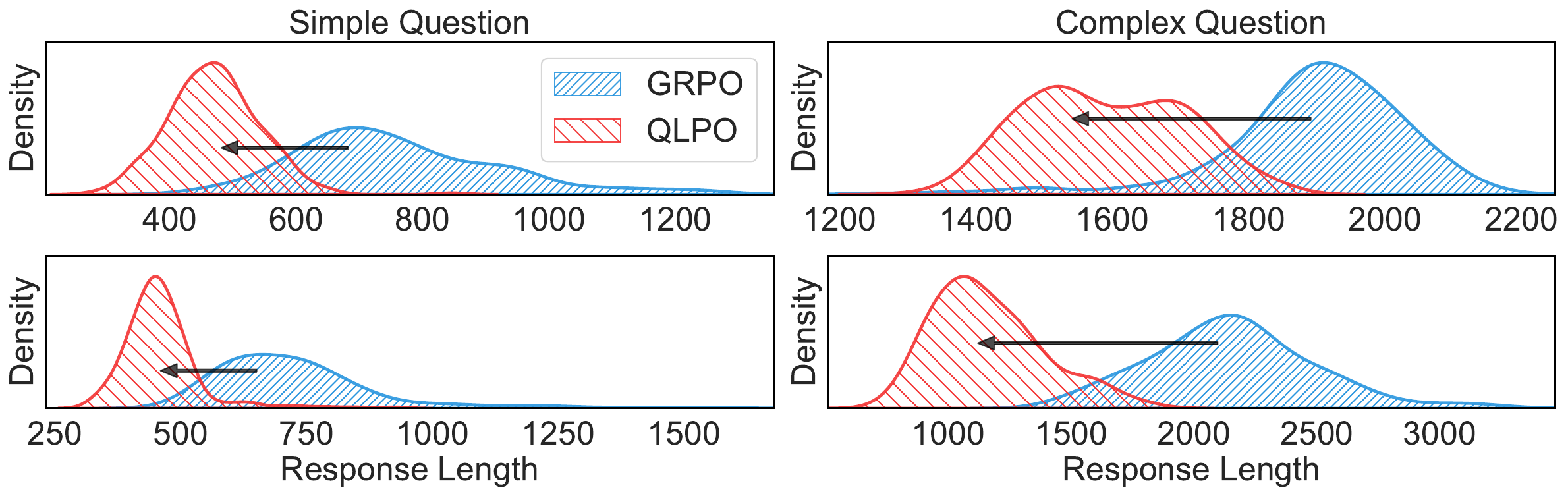}
    %\caption{Response Length Distribution Shift Across Question Complexity (DeepScaleR-1.5B-Preview, DAPO-MATH).}
    \caption{Response Length Distribution Shift Across Question Complexity for DeepScaleR-1.5B-Preview on GSM8K (left) and DAPO-MATH (right). QLPO shortens responses across both simple and complex questions, indicating a broad improvement in reasoning efficiency.}

    %\vspace{-12pt}
    \label{fig:diff-len}
\end{figure}

Across questions of varying difficulty, we observe comparable relative reductions in response length. As shown in Figure~\ref{fig:diff-len}, QLPO improves reasoning efficiency across different levels of problem difficulty. Quantitatively, the average length reduction is 30\% to 35\% on simpler benchmarks (GSM8K and MATH-500) and 33\% to 38\% on more complex benchmarks (OlympiadBench, GPQA and AIME24). This shows that QLPO promotes concise reasoning in a broad sense, rather than reflecting a compression effect confined to easier problems.

%% file: main-related.tex
\section{Related Work}
\textbf{RL-based Chain-of-Thought Compression.} A large recent literature explicitly treats CoT length as a controllable resource and optimizes an accuracy–cost objective, typically combining a task reward with a penalty proportional to generated reasoning tokens. Representative sequence-level approaches include direct length-penalty RL objectives and reward shaping that encourage concise correct solutions while preserving performance on harder instances \citep{wang2026darts,xiang2025alp,yuan2025lengthawareopt,su2025adlp}. Beyond global token penalties, several works introduce difficulty-adaptive schemes to realize “fast on easy, deep on hard” behavior \citep{shen2025dast,liang2025deepcompress,xiang2025alp,dumitru2025conciserl,singhal2024long}. More structured control has also been explored via step/segment-level early-exit or pruning formulations that more closely resemble optimal stopping, including serial early-exit rewards, step-aware cost definitions, and redundancy-aware penalties \citep{he2025smartthinker,wu2025steppruner,dai2025sgrpo,hong2025reconsidering}. In parallel, instance-level mode selection frameworks learn when to trigger explicit CoT using RL-trained controllers or control tokens, enabling large efficiency gains by avoiding unnecessary deliberation on easy queries \citep{tu2025autothink,lou2025adacot,deng2025hipo,zhang2025adr,liu2026laconic,yu2026stop,li2025leash,chen2025lspolengthawaredynamicsampling}.

\textbf{Sample Selection in LLM Training.} 
Sample selection is increasingly important in LLM training, especially in RL-style post-training with multiple responses per prompt. Some methods explicitly filter sampled responses before optimization: GFPO retains a top-\(k\) subset to favor token-efficient learning, while DLER uses dynamic sampling and batch filtering to avoid degenerate reward groups \citep{shrivastava2025gfpo,liu2025dler}. Other work emphasizes that rollout-group construction and length-aware sampling can themselves shape optimization behavior, as seen in large-scale RL systems such as DAPO and in dynamic sampling methods such as LSPO \citep{DAPO,chen2025lspolengthawaredynamicsampling}. Related compression pipelines further employ rejection sampling or selective optimization to focus learning on more useful responses or token segments \citep{rakotonirina2026ccc,he2025smartthinker,lou2025adacot,ha2025mera,li2025rorecomp}.

%% file: main-conclusion.tex
\section{Conclusion}
In this paper, we propose Quadrant-weighted sampling for Length-aware Policy Optimization (QLPO), a simple and effective reinforcement learning algorithm for mitigating length explosion in large reasoning models. By reshaping the training distribution through quadrant-weighted sampling, \name\ introduces an implicit preference for shorter reasoning without modifying the reward computation. Across a broad range of model scales and settings, \name\ consistently reduces response length by 30\% to 70\% while maintaining reasoning performance. Overall, \name\ provides a practical, robust, and easy-to-implement approach for improving the accuracy--efficiency trade-off of reasoning models.

\section*{Acknowledgment}
This work is supported by National Natural Science Foundation of China (U23B2048, 62402011), Fundamental and Interdisciplinary Disciplines Breakthrough Plan of the Ministry of Education of China (JYB2025XDXM108), High-performance Computing Platform of Peking University and The Fundamental Research Funds for the Central Universities. Xupeng Miao is the corresponding author.

%% file: appendix.tex
\section{Training Dynamics}
\label{append-dynamics}
Figure \ref{fig:append1} and Figure \ref{fig:append2} present the training curves of response length and accuracy in code-generation and multimodal settings, respectively. In both cases, QLPO shows the same qualitative pattern as in our main text experiments: compared with standard GRPO, it leads to shorter outputs over the course of training while preserving comparable task performance. The gap is especially clear in response length, whereas the accuracy curves remain close, suggesting that QLPO improves efficiency without introducing an obvious degradation in solution quality.
\begin{figure}[t]
    \centering
    \begin{subfigure}[t]{0.49\linewidth}
        \centering
        \includegraphics[width=\linewidth]{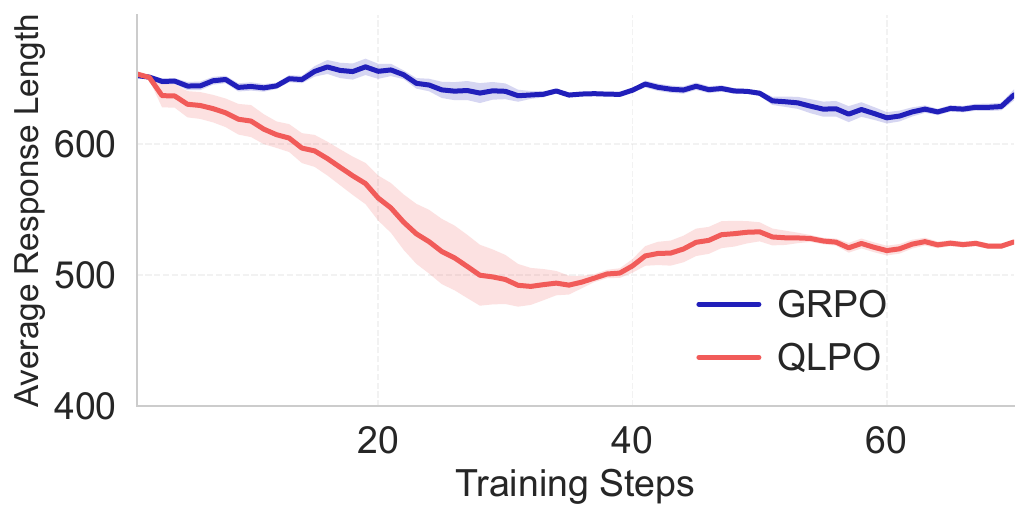}
        \label{fig:append1}
    \end{subfigure}
    \hfill
    \begin{subfigure}[t]{0.49\linewidth}
        \centering
        \includegraphics[width=\linewidth]{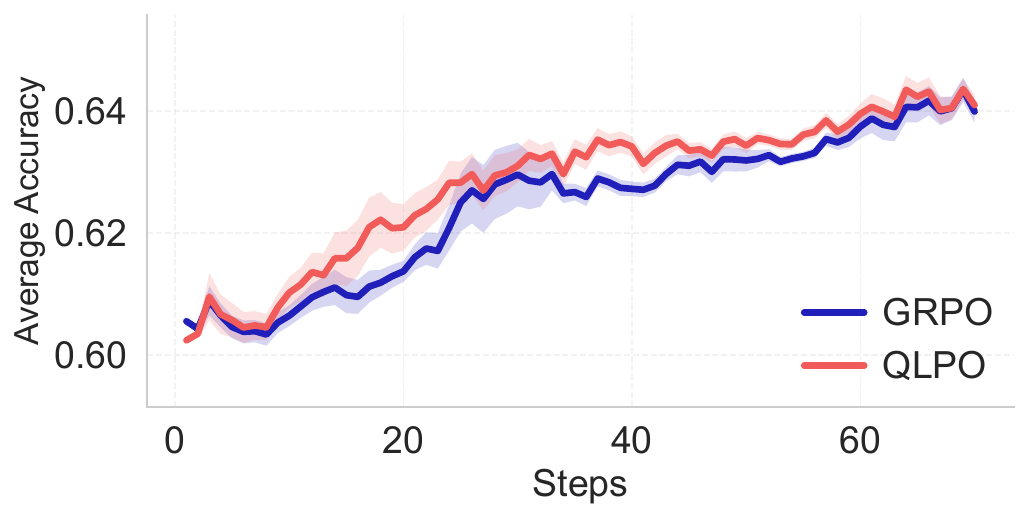}
        \label{fig:dist-append1}
    \end{subfigure}
    \caption{Phi-3.5-mini-instruct on Eurus-2-RL-Data.}
    \label{fig:append1}
\end{figure}

\begin{figure}[t]
    \centering
    \begin{subfigure}[t]{0.49\linewidth}
        \centering
        \includegraphics[width=\linewidth]{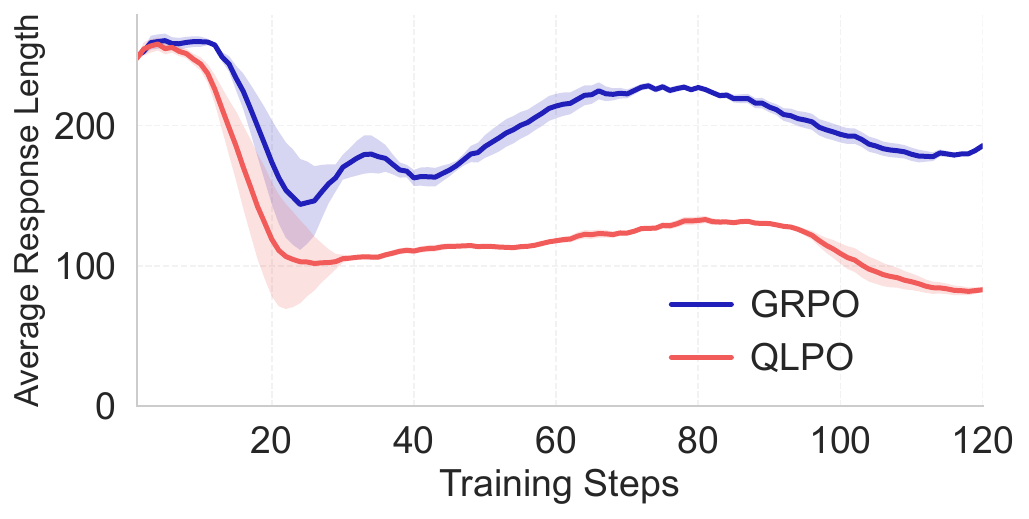}
        \label{fig:append2}
    \end{subfigure}
    \hfill
    \begin{subfigure}[t]{0.49\linewidth}
        \centering
        \includegraphics[width=\linewidth]{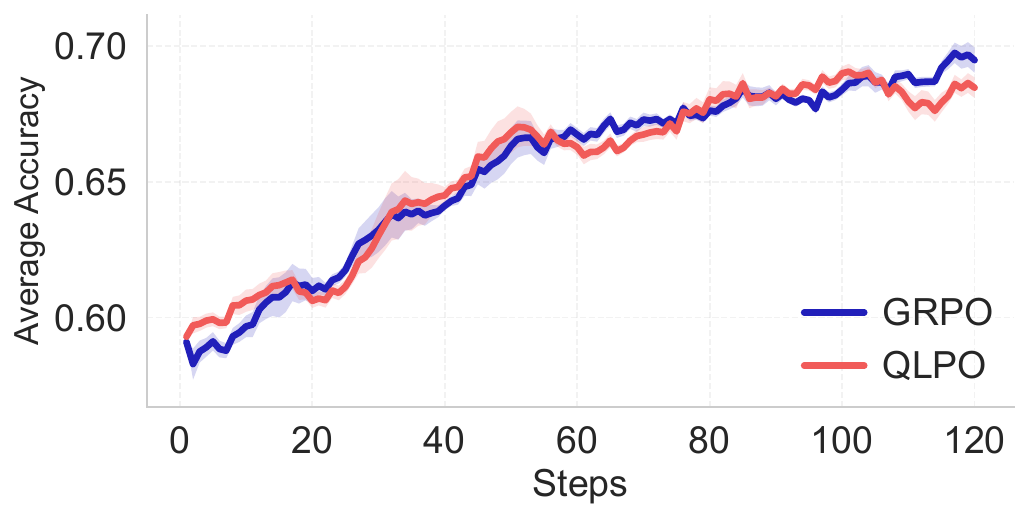}
        \label{fig:dist-append2}
    \end{subfigure}
    \caption{Qwen2.5-3B-VL on Geo3K.}
    \label{fig:append2}
\end{figure}
\section{Why Can QLPO Produce Shorter Responses with Higher Accuracy?}

To illustrate the qualitative difference induced by QLPO, we show a representative example from real model outputs below. The two responses are produced by Qwen2.5-32B models trained with GRPO and QLPO, respectively, on the prompt: ``Solve for \(x\): \(4(x+2)-3=2x+9\). Show your reasoning step by step and verify your final answer.''

\begin{quote}
\textbf{GRPO response:} ``To solve the equation \(4(x+2)-3=2x+9\) for \(x\), we follow these steps. First, distribute the 4 on the left-hand side to obtain \(4x+8-3=2x+9\). Next, combine like terms on the left-hand side, which gives \(4x+5=2x+9\). Then subtract \(2x\) from both sides to isolate the \(x\)-terms on one side, yielding \(2x+5=9\). After that, subtract 5 from both sides to get \(2x=4\). Finally, divide both sides by 2, so \(x=2\).

To verify the solution, substitute \(x=2\) back into the original equation: \(4(2+2)-3=2(2)+9\). Simplifying gives \(16-3=4+9\), and evaluating both sides yields \(13=13\). Since both sides are equal, the solution is verified. Therefore, the solution is 2.''

\textbf{QLPO response:} ``Distribute the 4 to get \(4x+8-3=2x+9\), then simplify to \(4x+5=2x+9\). Subtract \(2x\) from both sides to obtain \(2x+5=9\), subtract 5 to get \(2x=4\), and divide by 2 to find \(x=2\).

To verify, substitute \(x=2\) into the original equation: \(4(2+2)-3=2(2)+9\). This simplifies to \(16-3=4+9\), so \(13=13\). Therefore, the solution is correct.'' 
\end{quote}

Compared with the GRPO response, the QLPO response preserves the same overall reasoning structure---expanding and simplifying the equation, isolating the variable, and verifying the solution---but presents it more concisely and with less redundant narration. This example illustrates that the shorter response under QLPO need not arise from truncating essential reasoning. Instead, the compression mainly comes from removing repeated restatements, excessive verbal scaffolding, and other low-value tokens, while retaining the intermediate computations necessary for a correct solution.

This observation is consistent with a broader interpretation of why shorter responses could also be effective. CoT length is not equivalent to reasoning capability: a long response may reflect genuine multi-step reasoning, but it may also contain redundant explanations, repeated verification, or low-value detours that increase token count without contributing meaningful computation. Recent work has made this distinction more explicit. For example, \citet{chen2026thinkdeepjustlong} argue that surface-level token length is an unreliable proxy for reasoning quality, and suggests that effective reasoning is better characterized by ``deep-thinking tokens,'' i.e., tokens associated with substantial internal belief revision across model layers. From this perspective, QLPO does not simply bias the model toward shorter outputs; rather, by reallocating gradient signal away from inefficient trajectories and toward more informative ones, it may encourage the model to produce a higher concentration of tokens that correspond to genuinely useful reasoning. In this sense, QLPO may compress the surface form of CoT without reducing the effective depth of reasoning, allowing the model to preserve or even strengthen the critical intermediate computations that support accurate answers. This provides a plausible explanation for why shorter generations under QLPO need not hurt performance and can sometimes even improve it.

\section{Stability Across Random Seeds}
\label{sec:seed_stability}
To assess training stability, we repeat training with three random seeds on two representative settings: DeepSeek-R1-Distill-Qwen-7B and Qwen2.5-3B. For each seed, we evaluate the final checkpoint on the same five benchmarks used in Table~1. Accuracy and response length are first averaged over the five benchmarks, after which we report the mean and standard deviation across training seeds.
\begin{table}[h]
\centering
{
\begin{tabular}{lcccc}
\toprule
Model
& GRPO Acc.
& QLPO Acc.
& GRPO Len.
& QLPO Len. \\
\midrule
DS-7B
& \(68.7 \pm 0.5\)
& \(69.0 \pm 0.4\)
& \(9,256 \pm 1,076\)
& \(6,753 \pm 830\) \\
Qwen2.5-3B
& \(43.0 \pm 0.7\)
& \(43.4 \pm 0.6\)
& \(2,095 \pm 245\)
& \(1,071 \pm 182\) \\
\bottomrule
\end{tabular}
}
\caption{
Mean and standard deviation across three independent training seeds. Accuracy and response length are averaged over the five main benchmarks for each seed.
}
\label{tab:seed_stability}
\end{table}

Across both settings, the differences in average accuracy are small relative to seed-level variation, whereas the response-length reductions are substantially larger and consistent across seeds. QLPO reduces average length by 27.0\% on DS-7B and by 48.9\% on Qwen2.5-3B. These results indicate that the compression effect is not attributable to a single favorable random seed.

\section{Detailed Training-Time Results}
\label{app:training_time}

Section~\ref{sec:rollout_matched} presents the controlled
rollout-budget comparison on two representative models.
Here, we report the normalized wall-clock training time of
QLPO relative to standard 8-rollout GRPO across all five
model settings.

Table~\ref{tab:appendix_time_length} reports the normalized wall-clock training time of QLPO relative to standard 8-rollout GRPO across five models. Overall, QLPO introduces modest overhead in the evaluated single-turn reasoning settings. The normalized training time ranges from \(0.916\) to \(1.163\), corresponding to an \(8.4\%\) reduction to a \(16.3\%\) increase relative to
standard GRPO. Although QLPO samples twice as many initial candidates, the wall-clock increase is substantially smaller than the nominal increase in rollout count. This is because only the selected subset enters advantage estimation and back-propagation, and because QLPO progressively shortens generated responses during training. In some settings, such as Qwen2.5-32B, these savings offset the additional candidate generation and result in a lower
overall training time.

\begin{table}[h]
\centering
\small
\setlength{\tabcolsep}{8pt}
\begin{tabular}{lc}
\toprule
Model & QLPO time (GRPO \(=1.0\)) \\
\midrule
DeepScaleR-1.5B-Preview       & 1.035 \\
DeepSeek-R1-Distill-Qwen-7B   & 1.107 \\
Qwen2.5-3B                    & 1.132 \\
Qwen2.5-32B                   & 0.916 \\
Qwen3-30B-A3B                 & 1.163 \\
\bottomrule
\end{tabular}
\caption{
Normalized wall-clock training time of QLPO relative to standard GRPO with
\(K=M=8\). For each model, the training time of standard GRPO is normalized
to \(1.0\).
}
\label{tab:appendix_time_length}
\end{table}

\section{Full Experimental Configuration}
\label{app:full_configuration}

Table~\ref{tab:full_configuration} summarizes the training and evaluation
configuration used in the main experiments.

\begin{table}[h]
\centering
\small
\setlength{\tabcolsep}{6pt}
\begin{tabular}{ll}
\toprule
Hyperparameter & Value \\
\midrule
Maximum prompt length & 2k tokens \\
Maximum response length & 32k tokens \\
Prompt truncation & Left truncation \\
Base RL algorithm & GRPO \\
KL reward & Disabled \\
KL loss coefficient & 0.01 \\
Loss normalization & Token mean \\
Standard GRPO group size & \(8\) \\
QLPO candidate-pool size \(K\) & \(16\) \\
QLPO update-group size \(M\) & \(8\) \\
QLPO length coefficient \(\alpha\) & \(1/3\) \\
Rollout-matched GRPO group size & \(16\) \\
Optimizer & Adam \\
Learning rate & \(1\times10^{-7}\) \\
Warm-up steps & 10 \\
Weight decay & 0.1 \\
Gradient clipping & 1.0 \\
Prompt batch size & 128 \\
PPO mini-batch size & 256 \\
PPO clipping range & \([0.8,1.2]\) \\
Rollout temperature & 1.0 \\
Rollout top-\(p\) & 1.0 \\
Rollout top-\(k\) & \(-1\) \\
Evaluation temperature & 0.8 \\
Hardware & 32 NVIDIA H20 GPUs \\
Parallelism & FSDP + vLLM \\
Tensor-parallel size & 4 \\
Offloading & Enabled \\
Evaluation interval & Every 10 steps \\
Checkpoint interval & Every 50 steps \\
\bottomrule
\end{tabular}
\caption{Full configuration for the main experiments.}
\label{tab:full_configuration}
\end{table}

The maximum response length is 32k tokens. No more than 0.1\% of generated responses reach this
limit during training and evaluation. Therefore, the observed reductions in
response length are not primarily caused by output truncation.

\section{Limitations}
\label{sec:limitations}

While QLPO is an effective method for mitigating the length explosion problem in reasoning models, it has a few practical limitations. First, the method relies on an over-generation mechanism to construct the resampled training group. Although our experiments indicate that the training time often remains comparable to standard GRPO—since the model learns to generate shorter, faster-to-decode responses over time—sampling extra candidate trajectories can still incur additional computational overhead and potentially result in slower overall training, especially under constrained hardware configurations. Second, QLPO is explicitly designed for scenarios that suffer from excessively verbose chain-of-thought reasoning. Consequently, for models that naturally produce concise outputs, or for tasks that inherently require short, fixed-length text generation (e.g., simple question-answering without reasoning traces or rigid formatting tasks), applying QLPO provides little meaningful benefit. In such cases where the output length is already naturally bounded, standard reinforcement learning methods remain the more appropriate choice.

\section{Future Directions} QLPO is based on a quadrant-weighted sampling strategy, which is simple and effective. Future work could explore more fine-grained allocation rules that distinguish responses not only by binary correctness and relative length, but also by richer attributes such as difficulty, confidence, or step-level reasoning quality. Another promising direction is to replace the fixed resampling heuristic with adaptive policies that learn how much training weight should be assigned to different types of trajectories. Such extensions may yield better accuracy-efficiency trade-offs. However, such extensions might introduce additional complexity and warrant further exploration.